\documentclass[a4paper,fleqn]{cas-dc}

\usepackage[authoryear,longnamesfirst]{natbib}
\usepackage{bm}
\usepackage{amsmath, nccmath}
\usepackage{caption}
\usepackage{subcaption}

\makeatletter 
  \def\my@tag@font{\normalsize}
  \def\maketag@@@#1{\hbox{\m@th\normalfont\my@tag@font#1}}
  \let\amsmath@eqref\eqref
  \renewcommand\eqref[1]{{\let\my@tag@font\relax\amsmath@eqref{#1}}}
\makeatother

\newcommand{\mli}[1]{\mathrm{#1}} 

\def\tsc#1{\csdef{#1}{\textsc{\lowercase{#1}}\xspace}}
\tsc{WGM}
\tsc{QE}

\begin{document}
\let\WriteBookmarks\relax
\def\floatpagepagefraction{1}
\def\textpagefraction{.001}

\shorttitle{Predicting Ship Responses in Different Seaways using a Generalizable Force Correcting Machine Learning Method}

\shortauthors{K. E. Marlantes, P. J. Bandyk, and K. J. Maki}  

\title[mode = title]{Predicting Ship Responses in Different Seaways using a Generalizable Force Correcting Machine Learning Method}

\author[1]{Kyle E. Marlantes}[orcid=0000-0002-0003-3617]

\cormark[1]

\ead{kylemarl@umich.edu}

\credit{Conceptualization, Methodology, Software, Validation, Formal analysis, Investigation, Writing - Original Draft, Visualization}

\affiliation[1]{organization={University of Michigan, Department of Naval Architecture and Marine Engineering},
            city={Ann Arbor},
            state={MI},
            country={USA}}

\author[2]{Piotr J. Bandyk}

\credit{Methodology, Software, Resources, Writing - Review \& Editing, Data curation, Supervision}

\affiliation[2]{organization={MARIN, USA},
            city={Houston},
            state={TX},
            country={USA}}

\author[1]{Kevin J. Maki}

\credit{Conceptualization, Methodology, Writing - Review \& Editing, Supervision, Funding acquisition, Project administration}

\cortext[1]{Corresponding author}

\begin{abstract}
A machine learning (ML) method is generalizable if it can make predictions on inputs which differ from the training dataset. For predictions of wave-induced ship responses, generalizability is an important consideration if ML methods are to be useful in design evaluations. Furthermore, the size of the training dataset has a significant impact on the practicality of a method, especially when training data is generated using high-fidelity numerical tools which are expensive. This paper considers a hybrid machine learning method which corrects the force in a low-fidelity equation of motion. The method is applied to two different case studies: the nonlinear responses of a Duffing equation subject to irregular excitation, and high-fidelity heave and pitch response data of a Fast Displacement Ship (FDS) in head seas. The generalizability of the method is determined in both cases by making predictions of the response in irregular wave conditions that differ from those in the training dataset. The influence that low-fidelity physics-based terms in the hybrid model have on generalizability is also investigated. The predictions are compared to two benchmarks: a linear physics-based model and a data-driven LSTM model. It is found that the hybrid method offers an improvement in prediction accuracy and generalizability when trained on a small dataset.
\end{abstract}



\maketitle

\section{Introduction}\label{sect:intro}
Ocean-going ships encounter a large number of different seaways throughout their operational lifetimes. Therefore, to fully-characterize the performance and safety of a design, it is necessary to simulate the response of a vessel subject to a wide range of sea conditions. Unfortunately, this remains a challenge for high-fidelity physics-based methods, such as fully-nonlinear potential flow and Computational Fluid Dynamics (CFD) solvers, due primarily to computational cost, where the size of a mesh can grow to many millions of cells, and a simulation in a single seaway may take days or weeks to yield enough time-series for statistical inference. As a result, most industrial ship motion analyses are still performed with low-fidelity methods, such as linear potential flow, despite that the assumptions underlying linear methods are invalid in large seaways \citep{smith2017}. Viscous effects are also important for certain responses, such as roll and surge, and a growing interest in purely nonlinear phenomena, such as contemporary dynamic stability criteria \citep{spyrou2023}, further support the need for low-cost methods which capture the complex physics.

The computational cost of high-fidelity physics-based methods has led many researchers to develop data-driven surrogate models, which -- once trained -- can make time-series predictions at a fraction of the cost. Numerous examples of data-driven models for ship motion prediction and forecasting have been published, including \citep{xing2011}, \citep{xu2021}, \citep{silva2022}, \citep{liong2022}. The widespread adoption of Long Short-Term Memory (LSTM) neural networks \citep{hochreiter1997} for time-series modeling led to a proliferation of examples of their use in modeling marine dynamics \citep{xu2020}, \citep{liu2020}, \citep{xu2021}, \citep{guo2022}, and \citep{silva2022}, and wave forecasting \citep{meng2022}, \citep{kagemoto2022}, \citep{li2022}, \citep{yao2022}, \citep{breunung2023}. More recent work explores some alternative architectures, such as the slim-adversarial networks of \cite{geng2024} which utilize a recursive sliding window, transformer models \citep{zhangm2023}, embedding decomposition approaches \citep{diez2022b}, Gaussian process models \citep{rong2019}, and combined model approaches \citep{zhang2023ml}, \citep{zhangd2023}, \citep{li2024} which attempt to improve prediction accuracy beyond individual models. 

More recent research is shifting away from purely data-driven models, to hybrid -- or scientific -- machine learning (ML) methods which combine physics-based models with ML techniques, with a great deal of work outside of the marine dynamics realm already proposed, including the widely-adopted physics-informed neural networks (PINNs) of \cite{raissi2019}, Port-Hamiltonian neural networks \citep{desai2021}, and other approaches tailored to nonlinear dynamics \citep{pathak2018} or fluid dynamics \citep{pawar2021}. Works such as that of \cite{schirmann2022}, \cite{schirmann2023} point out the value of physics-based knowledge in improving the accuracy of data-driven ship motion predictions. Early work by \cite{weymouth2014} integrated physics-based knowledge as an input into data-driven models, \cite{wan2018} considered a hybrid methodology for extreme dynamics, and studies by \cite{skulstad2021}, \cite{wang2022}, \cite{nielsen2022}, and \cite{kanazawa2022} utilized hybrid approaches to predict ship trajectories for docking and maneuvering. \cite{diez2022} utilized a hybrid approach for ship motion forecasting and \cite{yang2022} proposed a hybrid method to predict the added resistance of ships sailing in waves. While the application of hybrid machine learning methods to ship hydromechanics is still relatively new compared to other data-driven approaches, research in other disciplines suggests that hybrid methods require less training data \citep{willard2020}, and may perform better than data-only methods when making predictions outside the original training dataset \citep{yang2022}.

\begin{table}
\caption{Training dataset size and consideration of different waves in the testing dataset for data-driven marine dynamics.}
\begin{tabular*}{\tblwidth}{L|R|C} 
\toprule
Study & \makecell{Training \\ dataset  \\ dimension$^{*}$ \\ $\times 10^{3}$} & \makecell{Test conditions \\ different \\ from training \\ conditions} \\ 
\midrule
\multicolumn{3}{C}{Data-only methods} \\
\midrule
\cite{xing2011} & 2.133 & No\\
\midrule
\cite{liu2020} & 1.875 & No\\
\midrule
\cite{xu2021} & 2.000 & No\\
\midrule
\cite{silva2022} & 15.360 & No\\
\midrule
\cite{liong2022} & 6.923 & No\\
\midrule
\cite{guo2022} & \textbf{1.858} & \textbf{Yes} \\
\midrule
\cite{zhangm2023} & \textbf{533.333} & \textbf{Yes}\\
\midrule
\cite{zhangd2023} & 0.111 & No \\
\midrule
\cite{li2024} & \textbf{1.750} & \textbf{Yes} \\
\midrule
\multicolumn{3}{C}{Hybrid methods} \\
\midrule
\cite{marlantes2022} & \textbf{0.147} & \textbf{Yes} \\
\midrule
\midrule
\textbf{Present Study} & \textbf{0.107} & \textbf{Yes} \\
\bottomrule
\multicolumn{3}{L}{\makecell{\footnotesize $^{*}$Training dataset dimension is the number of Zero-Up-Crossings \\ in the wave record: $N_{ZUC}$.}}
\end{tabular*}
\label{tab:sizes}
\end{table}

But if data-driven methods are considered from the eye of a ship designer, can they really help us design better ships? If the goal is to use data-driven methods to simulate the response of a prospective design in a large number of operating conditions, what properties must a data-driven model exhibit to make this a reality? When evaluating a novel design, real-world data does not exist. Furthermore, the quantity of high-fidelity simulation data which might be available will likely be limited, perhaps to only a few conditions. At a certain stage in the ship design process, model testing data might be available, but again this will likely be limited to a few seaways. To make a data-driven model especially useful for design evaluations, it is proposed that the model must possess two important characteristics: generalizability across sea conditions and good performance when using a small training dataset.

\textbf{Generalizability:} Domain generalizability is how well a model makes predictions subject to inputs which differ from the training dataset \citep{wangj2022}, \citep{kawaguchi2022}. For seakeeping predictions, domain generalizability concerns prediction accuracy in different wave conditions, different ship speeds, or different headings. In this work, only generalizability in wave conditions will be considered. To demonstrate generalizability in wave conditions, a machine learning method must be trained on response data from a seaway $S_{\mli{train}}$ with a significant wave height, $H_{s,\mli{train}}$ and wave period $T_{p,\mli{train}}$ and then the model must provide satisfactory predictions of responses in other seaways $S_{\mli{test}}$, which differ in $H_{s,\mli{test}}$ or $T_{p,\mli{test}}$. The degree to which $S_{\mli{test}}$ differ from $S_{\mli{train}}$, and the accuracy of the predictions when compared to reference data, determines the generalizability of a model. Generalizability increases the value of a model by leveraging a dataset to a wider range of applicability, which is critically important for design evaluations on novel designs.

\textbf{Size of the Training Data Set:} High-fidelity training data is costly, so it is desireable to have as small a training dataset as possible. The size of a training dataset can be measured in terms of time, the number of records of a fixed length (which is common in sequence-to-sequence methods), or the number of wave Zero-Up-Crossings, $N_{ZUC}$, which are present in a wave record. In this work, $N_{ZUC}$ is used as it roughly aligns with the number of response periods, and can be a helpful metric to inform the length of simulations. To determine how small a training dataset can be and still yield acceptable prediction accuracy, the $N_{ZUC}$ of a training dataset is systematically increased and predictions are compared to known reference data. Once successive prediction errors begin to converge within 5\%, for example, the minimum training dataset size is quantified. 

Despite the large number of papers exploring data-driven modeling for marine dynamics, few works consider the generalizability of a model to wave conditions that are not included in--or spanned by--the training dataset. Furthermore, the size of most training datasets is extremely large. As pointed out in \cite{portillojuan2022}, training datasets often include between 6,000 and 25,000 samples, with some studies exceeding 150,000 samples. To highlight this fact, a collection of studies are shown in Table~\ref{tab:sizes}, where the size of the training dataset, and whether or not model performance is evaluated in wave conditions that differ from the training dataset, are given. Only a single study, \cite{zhangm2023}, considered the generalizability of the trained model in wave conditions which differed distinctly from the training data, despite using an extremely large training dataset. A few additional studies considered test conditions which differed from the training data, however, the test conditions were contained within the parametric space of the training data, which does not strictly test generalizability. If the training dataset spans the wave conditions which are used for testing, the utility of the machine learning model is reduced more to that of an interpolant. This is perhaps why many studies suggest that machine learning methods are best suited to in-situ applications such as onboard monitoring and forecasting, digital twins \citep{schirmann2019}, and failure detection.

This paper further explores the neural-corrector method of \cite{marlantes2022}--a hybrid machine learning method which embeds a data-driven force correction term into a low-fidelity equation of motion--with specific focus on domain generalizability in irregular waves and the minimum size of the training dataset to ensure best performance. The method is applied to two case studies: first, the single-degree-of-freedom (DOF) responses of a nonlinear Duffing equation with irregular excitation, and second, the heave and pitch responses of a Fast Displacement Ship (FDS) in head seas, utilizing a multi-fidelity dataset generated using PANShip, a research code from the Maritime Research Institute Netherlands (MARIN). In both case studies, the testing dataset considers prediction performance in wave conditions which fall outside the training dataset. The influence of low-fidelity physics in the model is also considered. In addition, the performance of the method is benchmarked against a linear physics-based model and the predictions of a data-only LSTM method.

The paper is organized into four main sections. The remainder of Section \ref{sect:intro} will introduce the formulation of the neural-corrector method in Subsection \ref{sect:ncmethod} and the LSTM model in Subsection \ref{sect:lstm}. The performance metrics which are used to evaluate the quality of predictions are presented in Subsection \ref{sect:metrics}. Section \ref{sect:duff} gives results for the method when applied to responses from a nonlinear Duffing equation, with Subsection \ref{sect:duffsize} focusing on the size of the training data set, and Subsection \ref{sect:transduff} giving results on the generalizability in different significant wave heights. Section \ref{sect:fds} gives results for the Fast Displacement Ship (FDS) in irregular waves, first giving details on PANShip and the governing equations, with Subsection \ref{sect:singlesea} giving details on the model configuration and training dataset size, and Subsection \ref{sect:transfds} demonstrating the generalizability of the methods in irregular wave conditions which differ in wave height and peak period. Section \ref{sect:concl} briefly summarizes the work and offers conclusions.

\subsection{Neural-Corrector Method}\label{sect:ncmethod}

The neural-corrector method is a hybrid machine learning method that relates a high-fidelity and low-fidelity model by a force correction that is modeled using an artificial neural network \citep{marlantes2022}. To illustrate, Eq. \eqref{eq:heqm} is the high-fidelity model, indicated by the superscript $(h)$, and the solution to this ordinary differential equation is the high-fidelity state $\mathbf{\xi}^{(h)}$~=~$\{\ddot{\xi},\dot{xi},\xi\}$. Eq. \eqref{eq:leqm} is the low-fidelity model--a model that is inexpensive to solve but lacks accuracy--indicated by superscript $(l)$, where the solution to the equation is the low-fidelity state $\ddot{\xi}^{(l)}$.

\begin{align}
m\ddot{\xi}^{(h)} &= f^{(h)} \label{eq:heqm} \\
m\ddot{\xi}^{(l)} &= f^{(l)} \label{eq:leqm} 
\end{align}

Adding the low-fidelity force model $f^{(l)}$ from Eq. \eqref{eq:leqm} to the right-hand-side of Eq. \eqref{eq:heqm} results in a force correction term $\delta$, as shown by Eq. \eqref{eq:nceqm}, where the force correction is the difference between $f^{(h)}$ and $f^{(l)}$. 

\begin{align}
m\ddot{\xi}^{(h)} &= f^{(l)} + \delta \label{eq:nceqm}
\end{align}

An analytical model for $\delta$ may not be available, so it is modeled using an artificial neural network, which introduces an error $\epsilon~=~\delta~-~\delta ^{*}$, where $\delta^{*}$ is the approximate force correction obtained by the trained model. Considering this error, Eq. \eqref{eq:nceqm} becomes Eq. \eqref{eq:nceqm2}. A solution to Eq. \eqref{eq:nceqm2} will yield an approximate high-fidelity state $\ddot{\xi}^{*}$ which will approach $\ddot{\xi}^{(h)}$ as $\epsilon \to 0$.

\begin{equation}
m\ddot{\xi}^{*} = f^{(l)} + \delta^{*} \label{eq:nceqm2}
\end{equation}

Both recurrent neural networks like Long Short-Term Memory (LSTM) and simpler feed-forward densely-connected multi-layer networks have been used to model $\delta$, but it is found in \cite{marlantes2023} that relatively small, simple networks are sufficient for ship hydrodynamics problems when constant in terms of correction, with the added benefit that they are inexpensive to train and evaluate. The neural network configuration used in this paper is shown in Table \ref{tab:ncml}. The primary consideration when designing the network is to accomodate numerical integration of Eq. \eqref{eq:nceqm2}. To this end, $\delta$ is modeled as a function of $k$-length discrete sequences of prior state $\{\xi\}_{n-k-1}^{n}$, $\{\dot{\xi}\}_{n-k-1}^{n}$, $\{\ddot{\xi}\}_{n-k-1}^{n}$ and the wave elevation $\{\eta\}_{n-k-1}^{n}$, where the current time is $t^{n}$. Therefore, the state and wave elevation comprise the input features of the neural network and the output is simply $\delta^{*,n+1}$.

\begin{table}
\caption{Configuration of neural-corrector ML model.}
\begin{tabular*}{\tblwidth}{R|R|R|R|R} 
\toprule
Input & Output & Hid'n L's & Cells & Cells/Layer \\
\midrule
$\ddot{\xi}$, $\dot{\xi}$, $\xi$, $\eta$ & $\delta_{n+1}$ & 2 & ReLu & 30 \\
\bottomrule
\multicolumn{5}{L}{\footnotesize Input features are discrete sequences from $t_{n-k-1} ... t_{n}$.} \\
\multicolumn{5}{L}{\footnotesize Total number of trainable parameters: 7,621}
\end{tabular*}
\label{tab:ncml}
\end{table}

\subsection{Long Short-Term Memory (LSTM)}\label{sect:lstm}

Long Short-Term Memory (LSTM) neural networks are popular machine learning architecture for data-only time series predictions of ship responses in waves. The work of \cite{xu2020} gives a good overview of their application to the problem, and many authors have offered similar works. The models can be used in a sequence-to-sequence (S2S) approach, where a length of wave elevation $\eta$ time series serves as input to the model, and the output is the time series of ship responses (position $\xi$, velocity $\dot{\xi}$, or acceleration $\ddot{\xi}$) of equivalent length. 

\begin{table}
\caption{Configuration of data-only S2S-LSTM model.}\label{doconfig}
\begin{tabular*}{\tblwidth}{R|R|R|R|R} 
\toprule
Input & Output & Hidden Layers & Cells & Cells/Layer \\
\midrule
$\eta$ & $\ddot{\xi}$, $\dot{\xi}$, $\xi$ & 3 & LSTM & 50 \\
\bottomrule
\multicolumn{5}{L}{\footnotesize Input features are discrete sequences from $0...t_{N}$.} \\
\multicolumn{5}{L}{\footnotesize Total number of trainable parameters: 71,153}
\end{tabular*}
\label{tab:lstmml}
\end{table}

LSTM neural networks are particularly successful in S2S time series modeling because they retain knowledge of a previous state. The networks are composed of multiple cells, much like the nodes of a traditional feed-forward neural network, however, each cell encompasses an update, forget, and output gate, instead of a single activation function. The flow of information between cells is often described using a conveyor belt analogy, where the conveyor carries information from cell-to-cell and the gates in each cell add or remove information from the conveyor throughout the sequence \citep{colah2015}. Thus, the model retains a form of memory. To train the model, the gates are optimized via a learning algorithm--a popular choice is the Adam optimizer from \cite{kingma2015}. The number of layers, and the number of cells per layer, are configurable, and there is no strict guidance on the configuration, though \cite{vanhoudt2020} and \cite{portillojuan2022} indicate that more complex networks may not learn well.\par

Given the popularity of data-only LSTM methods for time series modeling in the literature, they are included in this study as a benchmark. The model is configured according to Table \ref{tab:lstmml}, which matches the LSTM models used by \cite{xu2021}. 

\subsection{Performance Metrics}\label{sect:metrics}

The average $L_{2}$ error, $L_{\infty}$ error, given by Eqs. \eqref{eq:l2} and \eqref{eq:linf}, respectively, are used to evaluate the accuracy of the time series predictions $\hat{x}$ to that of the reference $x$ in terms of RMS and extreme values. 

\begin{equation}
L_{2} = \sqrt{ \frac{\sum_{i}^{N} (\hat{x} - x)^{2}}{N}} \label{eq:l2}
\end{equation}

\begin{equation}
L_{\infty} = \max (|\hat{x} - x|) \label{eq:linf}
\end{equation}

However, such measures are sensitive to small phase errors. As a more powerful measure of performance, the Jensen-Shannon divergence ($\mli{JSD}$), given by Eq. \eqref{eq:jsd}, is used to estimate the entropy of the predicted response pdf $Q$ relative to a known reference pdf $P$, where $M$ is the mixture of the two \citep{LIN1991}. The Jensen-Shannon divergence is based on the Kullback-Leibler divergence $D$, given by Eq. \eqref{eq:kld}, which is a measure of the relative entropy between the model distribution $K$ and the reference distribution $M$, both defined over the domain $\chi$ \citep{kullback1951}. It can be thought of as a measure of information loss, or expected surprise, if a certain distribution is used to model a reference distribution. A lower $\mli{JSD}$ means the model is closer to the reference, with a divergence of zero meaning the two distributions are identical.

\begin{gather}
\mathrm{JSD}(P\|Q) = \frac{1}{2}D(P\|M) + \frac{1}{2}D(Q\|M)  \label{eq:jsd} \\
M = \frac{1}{2}(P+Q) \notag \\
D(K\|M) = \sum_{x \in \chi} K(x) \log \left( \frac{K(x)}{M(x)} \right) \label{eq:kld}
\end{gather}

The $\mli{JSD}$ is especially useful for evaluating the generalizability of a model subject to variations in excitation as it is sensitive deviations in the tails of the response distributions, which give a good indication of the accuracy of extreme values. Furthermore, it is possible for a model to yield favorable $L_{2}$ errors, yet provide poor predictions of the tails of the distribution, or alternatively a model may yield unfavorable $L_{\infty}$ errors, yet provide favorable predictions of the distribution if the errors are not focused in the tails. Accurate predictions of the response distribution is especially important for marine dynamics applications, where the predicted pdf will be used to select design values. 

\section{Duffing Equation with Nonlinear Excitation}\label{sect:duff}

Eq. \eqref{eq:nceqm} depends on $f^{(l)}$, the choice of which will determine the physics that are learned by the ML model for $\delta$ and those that are modeled analytically. This section investigates the influence $f^{(l)}$ has on the performance of the model, specifically any implications on generalizability. A single degree-of-freedom (DOF) forced Duffing equation is used as a theoretical model as it is a popular model of the nonlinear hydrodynamics problem of a ship rolling in waves. The Duffing equation used in this work is given by Eq. \eqref{eq:duffing}, where $c_{1}$ and $c_{3}$ are the linear and cubic hydrostatic restoring coefficients, $b_{1}$ and $b_{2}$ are the linear and quadratic hydrodynamic damping coefficients, and the wave excitation forcing due to irregular waves is expressed as a summation of harmonic wave components multiplied by a forcing coefficient $\beta$ which has units of force per length. The wave excitation is made nonlinear by including the state $z$ in the amplitude, modified by a coefficient $\alpha$.

\begin{equation}
m\ddot{z} = \sum_{i} \beta(\zeta_{i}-\alpha z)\cos(\omega_{i}t+\phi_{i}) - c_{1}z - c_{3}z^{3} - b_{1}\dot{z} - b_{2}|\dot{z}|\dot{z} \label{eq:duffing}
\end{equation}

The wave component angular frequencies $\omega_{i}$ and amplitudes $\zeta_{i}=\sqrt{2S(\omega_{i})\delta\omega}$, where $\delta\omega$ is the sample bandwidth, are sampled from a generic wave energy spectrum $S(\omega)$ given by Eq. \eqref{eq:Sw}, where $H_{s}$ is the significant wave height and $\omega_{p}$ is the peak frequency. The component phase angles $\phi_{i}$ are selected randomly from the range $[-2\pi:2\pi]$.

\begin{equation}
S(\omega) = H_{s}^{2}\frac{5}{3}\frac{\omega_{p}^{4}}{\omega^{5}} \exp\left(-\frac{5}{4}\left(\frac{\omega_{p}}{\omega}\right)^{4}\right) \label{eq:Sw}
\end{equation}

Five different low-fidelity forcing models are proposed, given by Eqs. \eqref{eq:fA}-\eqref{eq:fE}, with Eq. \eqref{eq:fA} having the most physics retained (and consequently the least physics that must be learned in $\delta$). In Eq. \eqref{eq:fE}, the entire forcing function must be learned by the ML model.

\begin{align}
f_{A}^{(l)} &= \sum_{i} \zeta_{i}\cos(\omega_{i}t+\phi_{i}) - c_{1}z^{(h)} - b_{1}\dot{z}^{(h)} \label{eq:fA}\\
f_{B}^{(l)} &= - c_{1}z^{(h)} - b_{1}\dot{z}^{(h)} \label{eq:fB}\\
f_{C}^{(l)} &= - c_{1}z^{(h)} \label{eq:fC}\\
f_{D}^{(l)} &= \sum_{i} \zeta_{i}\cos(\omega_{i}t+\phi_{i}) - c_{1}z^{(h)} \label{eq:fD}\\
f_{E}^{(l)} &= 0 \label{eq:fE}
\end{align}

Table \ref{tab:dufffl} also shows the force components that are retained as physics and modeled analytically and those that are data-driven and modeled using ML for each choice of forcing model $f^{(l)}$. The models are ordered according to their eigenvalues, where models A / B, and C / D, share the same eigenvalues, respectively, and E has a unique set of eigenvalues. Appendix \ref{app:eig} includes details about how the choice of $f^{(l)}$ changes the eigenvalues of Eq. \eqref{eq:nceqm} at the limit when $\delta \to 0$.

\begin{table}
\caption{Forces retained as physics (\textbf{P}) in $f^{(l)}$ or learned by ML in $\delta$ for the Duffing equation.}\label{tab:dufffl}
\begin{tabular*}{\tblwidth}{|L|c|c|c|c|c|} 
\toprule
 & \multicolumn{5}{c|}{Forcing Model } \\
\midrule
Force Description & A & B & C & D & E\\ 
\midrule
Linear Restoring & \textbf{P} & \textbf{P} & \textbf{P} & \textbf{P} & ML\\
Nonlinear Restoring & ML & ML & ML & ML & ML\\
Linear Damping & \textbf{P} & \textbf{P} & ML & ML & ML\\
Nonlinear Damping & ML& ML & ML & ML & ML\\
Linear Excitation & \textbf{P} & ML & ML & \textbf{P} & ML\\
Nonlinear Excitation & ML & ML & ML & ML & ML\\
\bottomrule
\end{tabular*}
\end{table}

The Duffing equation is configured using $m$~=~$1.0$, $c_{1}$~=~$1.0$, $c_{3}$~=~$0.01$, $b_{1}$~=~$0.1$, $\beta$~=~1.0, $b_{2}$~=~$\alpha$~=~$0$ so that the only nonlinear term is the cubic restoring force. A direct numerical solution of Eq. \eqref{eq:duffing} yields the high-fidelity responses $z^{(h)}$, $\dot{z}^{(h)}$, $\ddot{z}^{(h)}$.

\subsection{Size of Training Dataset}\label{sect:duffsize}

\begin{figure}
	\centering
		\includegraphics{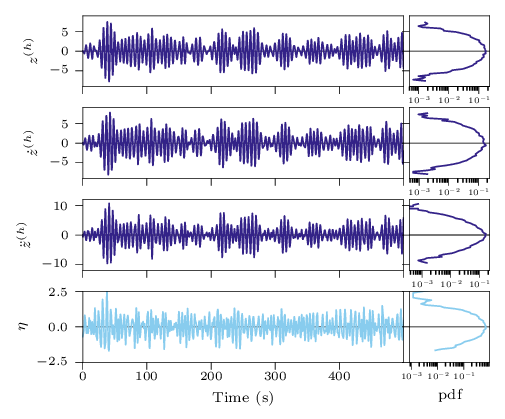}
	  \caption{An example of high-fidelity responses $z^{(h)}$, $\dot{z}^{(h)}$, $\ddot{z}^{(h)}$ computed using the Duffing equation for use as training data. The wave elevation $\eta$ comes from Eq. \eqref{eq:Sw} where $H_{s}$~=~1.0~m, $\omega_{p}$~=~1.0~rad/s, and the total number of Zero-Up-Crossings (ZUC) is 108.}\label{fig:dufftrain}
\end{figure}

When training data is expensive, it is desired that a model use the smallest training dataset possible, so the influence of training dataset size on prediction accuracy is first investigated. Using a significant wave height $H_{s}$~=~1.0~m and a peak frequency of $\omega_{p}$~=~1.0~rad/s, irregular wave records of $\eta$ ranging in total length of 10~s up to 1000~s are generated, and the corresponding high-fidelity responses $z^{(h)}$, $\dot{z}^{(h)}$, $\ddot{z}^{(h)}$. A time step of $\Delta t$~=~0.1~s is used and the frequency bandwidth $\delta\omega$ when sampling Eq. \eqref{eq:Sw} is taken such that the repeat period of the resulting summation is equal to the length of the time series. The responses and wave realization are shown in Figure \ref{fig:dufftrain}. 

Using the data in Figure \ref{fig:dufftrain}, the force correction $\delta$ is computed for each low-fidelity forcing model in Table \ref{tab:dufffl} and an ML model described by Table \ref{tab:ncml} is trained for each configuration. A stencil length $k$~=~5 is used, per the recommendations outlined in \cite{marlantes2023}. A small $k$ is effective in this case because the nonlinear force components are functions only of the instantaneous state variables $z$ and $\dot{z}$.

\begin{figure}
	\centering
		\includegraphics{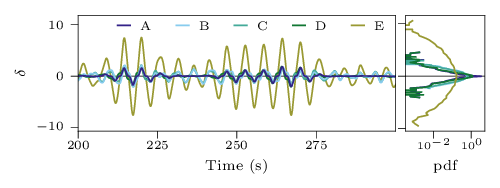}
	  \caption{An example of the force correction $\delta$ computed using different low-fidelity forcing models $f^{(l)}$ for use as training data, waves are $H_{s}$~=~1.0~m, $\omega_{p}$~=~1.0~rad/s. For clarity, only time series from 200~s to 300~s is shown, but the pdfs encompass the entire 500~s time series.}\label{fig:duffdeltatrain}
\end{figure}

\begin{figure}
	\centering
		\includegraphics{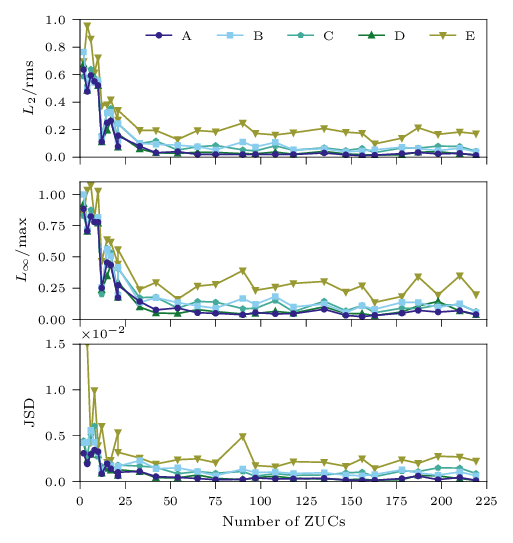}
	  \caption{$L_{2}$, $L_{\infty}$, and $\mli{JSD}$ prediction errors vs training data set size for each of the five low-fidelity forcing models $f^{(l)}$ in Table \ref{tab:dufffl}. The training data set is measured in terms of wave Zero-Up-Crossings (ZUCs). The testing waves come from Eq. \eqref{eq:Sw} where $H_{s}$~=~1.0~m, $\omega_{p}$~=~1.0~rad/s, and responses are solved using $\Delta t$~=~0.1~s with a stencil length $k$~=~5.}\label{fig:trainsize}
\end{figure}

The trained models are used to make predictions of the response in 1000~s of irregular waves with the same $H_{s}$~=~1.0~m and $\omega_{p}$~=~1.0~rad/s, but with different random phase angles. For each model, the $L_{2}$, $L_{\infty}$, and $\mli{JSD}$ are computed. Figure \ref{fig:trainsize} shows the prediction errors vs training dataset size for each low-fidelity forcing model. Note that the training dataset size is given as the number of Zero-Up-Crossings (ZUCs), $N_{ZUC}$, in the wave record, as this is a more meaningful measure of response encounters than time alone.

Figure \ref{fig:trainsize} shows that prediction errors of the five different low-fidelity forcing models converge at roughly the same rate relative to the size of the training dataset. Datasets of approximately $N_{ZUC}$~$\geq$~50 yield similar prediction errors.

\subsection{Generalizability in $H_{s}$}\label{sect:transduff}

\begin{figure}
	\centering
		\includegraphics{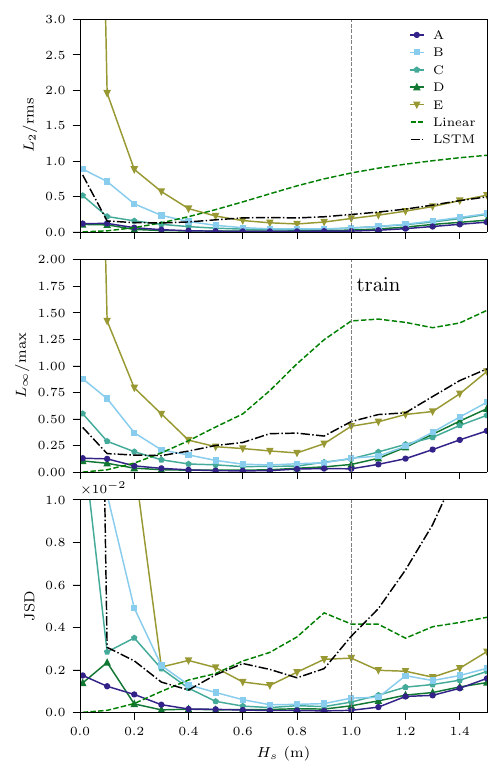}
	  \caption{$L_{2}$, $L_{\infty}$, and $\mli{JSD}$ prediction errors vs $H_{s}$ for each of the five low-fidelity forcing models $f^{(l)}$ in Table \ref{tab:dufffl}. The training wave condition is shown with a vertical dashed line corresponding to $H_{s}$~=~1~m.}\label{fig:errhs}
\end{figure}

A testing dataset is generated for a range of significant wave heights $H_{s}$ from 0.01~m to 1.5~m and a peak frequency $\omega_{p}$~=~1.0~rad/s over 1000~s of time with $\Delta t$~=~0.1~s. The component phase angles $\phi_{i}$ are selected randomly. The ML models for each of the five low-fidelity forcing models from Table~\ref{tab:dufffl} for $N_{ZUC}$~=~100 and $H_{s}$~=~1.0~m are used to make predictions of the responses in the range of $H_{s}$ from the testing dataset. The $L_{2}$, $L_{\infty}$, and $\mli{JSD}$ metrics for the predictions are computed over the last 900~s of time series, omitting the first 100~s as it is a transient region. Figure~\ref{fig:errhs} shows the performance metrics for each low-fidelity forcing model, the LSTM predictions, and a benchmark linear model--model A from Table~\ref{tab:dufffl} without the ML parts--over the range of test significant wave heights.

\begin{figure}
	\centering
		\includegraphics{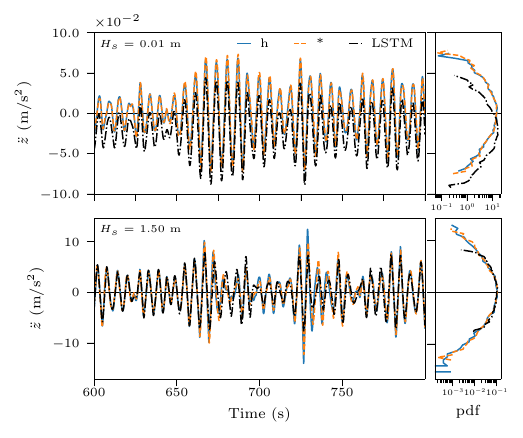}
	  \caption{Predicted responses for proposed model $*$ and LSTM model compared to the target high-fidelity data $h$ in the smallest $H_{s}$~=~0.01~m and largest $H_{s}$~=~1.5~m waves in the testing data set. Training waves: $H_{s}$~=~1.0~m. Only 200~s of time series are shown for clarity, however, the pdfs include the entire 900~s of data.}\label{fig:tsduffing}
\end{figure}

Figure \ref{fig:errhs} shows model A--the model that retains the most physics as analytical terms--performs better than the other models, and the LSTM benchmark, in nearly all wave conditions. This suggests that retaining more physics in the low-fidelity forcing model $f^{(l)}$ improves the generalizability of the overall model. This is especially evident at small wave heights, where the low-fidelity physics enforce the correct dynamics at the linear limit. This behavior can be explained analytically by considering the eigenvalues of the system when $\delta \to 0$, the details for which are found in Appendix~\ref{app:eig}. The $L_{2}$ and $L_{\infty}$ errors from the LSTM predictions are roughly in line with model E, which is perhaps intuitive as it is almost a purely data-driven model as well. However, the LSTM model outperforms model E at low wave heights.

The $\mli{JSD}$ errors diverge considerably at wave heights less than 0.5~m, where all but models A and D stuggle to converge to the linear limit. This is primarily due to errors in the tails of the predicted distributions. In other words, the other models, especially the LSTM, appear to yield satisfactory performance when evaluated using mean values of the response, but fail to predict the entire distribution when tested on waves outside the original training data set. This behavior is more clearly observed in Figure~\ref{fig:tsduffing}, which shows the time series and pdfs of the predictions using forcing model A and the LSTM benchmark compared to the high-fidelity reference data $(h)$ for the smallest $H_{s}$~=~0.01~m and largest $H_{s}$~=~1.5~m significant wave heights.

\section{Responses of a Fast Displacement Ship in Head Seas}\label{sect:fds}

\begin{figure}
	\centering
		\includegraphics{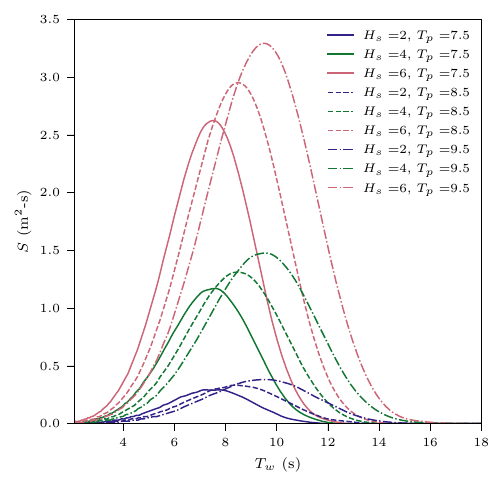}
	  \caption{JONSWAP wave spectra.}\label{fig:specs}
\end{figure}

This section considers the responses of a Fast Displacement Ship (FDS) in irregular waves. The FDS is a semi-displacement hull with a length of 100~m, a breadth of 12.5~m, and a displaccement of 1,608~metric tons. Table~\ref{tab:fdspart} gives the particulars of the vessel as they pertain to this work. Nine irregular head seas are prepared using a JONSWAP spectrum with significant wave heights ranging from $H_{s}$~=~$2.0,4.0,6.0$~m and peak periods $T_{p}$~=~$7.5,8.5,9.5$~s with a peak shape factor of 1.0. Figure~\ref{fig:specs} shows the wave spectra.

\begin{table}
\caption{Particulars of the Fast Displacement Ship (FDS).}\label{tab:fdspart}
\begin{tabular}{|c|R|L|} 
\toprule
$L_{PP}$ (m) & 100 & Length between perpendiculars \\
$L_{WL}$ (m) & 99.982 & Length on waterline \\
$B$ (m) & 12.502 & Breadth moulded on WL \\
$T_{F}$ (m)&  3.125 & Draft moulded on FP (from BL) \\
$T_{A}$ (m)&  3.125 & Draft moulded on AP (from BL) \\
$\nabla$ (m$^{3}$)&  1568.4 & Displacement volume moulded \\
$\Delta$ (t)&  1607.6 & Displacement mass in seawater \\
$K_{yy}$ (m)&  25 & Radius of gyration around $y$-axis\\
$V$ (knots)& 35.4 & Ship speed \\
$Fr$ (-)& 0.58 & Froude number (length based) \\
$\mu$ (deg) & 180 (head) & Wave heading relative to ship \\
\bottomrule
\end{tabular}
\end{table}

PANShip \citep{vanwalree2002}, \citep{vanwalree1999}, a research code from the Maritime Research Institute Netherlands (MARIN), is used to compute the responses of the FDS in the nine wave conditions. The code is a three-dimensional boundary element method (BEM) based on potential flow theory and yields predictions at different levels of fidelity by adjusting the method by which different forcing components are computed. In this work, two levels are considered, as shown in Table~\ref{tab:pannl}, to match the approach described in Section~\ref{sect:ncmethod}. The low-fidelity $(l)$ results are first-order, time domain responses computed using the frequency-domain impulse response function (IRF) approach and linearized excitation and restoring forces. The high-fidelity $(h)$ results use a body-exact formulation with a direct pressure integration (DPI) over the surface of the hull to compute the second-order hydrodynamic forces. The cost of the nonlinear computations is on the order of 10-100x real-time.

\begin{table}
\caption{Levels of fidelity in PANShip.}\label{tab:pannl}
\begin{tabular*}{\tblwidth}{|c|R|R|} 
\toprule
Force & Low-fidelity $(l)$ & High-fidelity $(h)$ \\
\midrule
Hydrostatic & Linear & Body-exact \\
Froude-Krylov & Linear & Body-exact \\
Radiation/diffraction & IRF & Pressure \\
Second-order & no & yes \\
\bottomrule
\end{tabular*}
\end{table}

Responses in PANShip are solved in a right-handed, ship-fixed reference frame where the longitudinal coordinate $x_{s}$ is positive forward, the transverse coordinate $y_{s}$ is positive to port, and the vertical coordinate $z_{s}$ is positive upward, with the origin at the center of gravity. This choice diagonalizes the physical mass matrix and also makes the equations easier to solve numerically. Eqs. \eqref{eq:transpan} and \eqref{eq:rotpan} give the governing equations of motion for a vessel free to move in translation (surge, sway, heave) and rotation (roll, pitch, yaw), respectively, where $\mathbf{m}$ is the 3x3 physical mass matrix, $\mathbf{I}$ is the 3x3 physical rotational inertia matrix, $\mathbf{u}$ is the ship-fixed velocity vector in translation, $\boldsymbol{\omega}$ is the ship-fixed velocity vector in rotation, and $\mathbf{f}$ and $\boldsymbol{\tau}$ are 3x1 vectors of ship-fixed external forces and moments, including the hydrodynamic added mass and damping, and hydrostatic restoring terms.

\begin{align}
\mathbf{m}\cdot(\mathbf{\dot{u}}+\boldsymbol{\omega}\times\mathbf{u}) &= \mathbf{f} \label{eq:transpan} \\
\mathbf{I}\cdot\boldsymbol{\dot{\omega}}+\boldsymbol{\omega}\times\mathbf{I}\cdot\boldsymbol{\omega} &= \boldsymbol{\tau} \label{eq:rotpan}
\end{align}

Following the formulation in Section \ref{sect:ncmethod} yields Eqs. \eqref{eq:traneomcorr} and \eqref{eq:roteomcorr}, where the ML-driven force and moment correction $\boldsymbol{\delta}$ is shown split into the translational $\boldsymbol{\delta}_{f}$ and rotational $\boldsymbol{\delta}_{\tau}$ components, each 3x1 vectors respectively.

\begin{align}
\mathbf{m}(\mathbf{\dot{u}}^{(h)}+\boldsymbol{\omega}^{(h)}\times\mathbf{u}^{(h)}) &= \mathbf{f}^{(l)} + \boldsymbol{\delta}_{f} \label{eq:traneomcorr} \\
\mathbf{I}\boldsymbol{\dot{\omega}}^{(h)}+\boldsymbol{\omega}^{(h)}\times\mathbf{I}\cdot\boldsymbol{\omega}^{(h)} &= \boldsymbol{\tau}^{(l)} + \boldsymbol{\delta}_{\tau} \label{eq:roteomcorr}
\end{align}

The low-order force and moment models $\mathbf{f}^{(l)}$ and $\boldsymbol{\tau}^{(l)}$ are selected in accordance with the findings in Section \ref{sect:duff}, where it is shown that generalizability is improved when additional low-fidelity terms are included. Eq. \eqref{eq:flpan} gives the low-order forcing model used in this section, where the linear wave excitation $\mathbf{f}^{(l)}_{w}$, along with the hydrodynamic added mass and damping, and hydrostatic restoring forces and moments are retained. Note that the restoring forces must be computed from the earth-fixed positions $\mathbf{x}$ and Euler-angles $\boldsymbol{\phi}$, which are obtained by applying a transformation to the ship-fixed velocities $\mathbf{u}$ and $\boldsymbol{\omega}$ and integrating. Details about the implementation are found in Appendix \ref{app:impl}.

\begin{equation}
\begin{Bmatrix}\mathbf{f}^{(l)} \\ \boldsymbol{\tau}^{(l)} \end{Bmatrix} = \mathbf{f}^{(l)}_{w} -\mathbf{A}\begin{Bmatrix} \mathbf{\dot{u}}^{(h)} \\ \boldsymbol{\dot{\omega}}^{(h)} \end{Bmatrix} - \mathbf{B}\begin{Bmatrix} \mathbf{u}^{(h)} \\ \boldsymbol{\omega}^{(h)} \end{Bmatrix} - \mathbf{C}\begin{Bmatrix} \mathbf{x}^{(h)} \\ \boldsymbol{\phi}^{(h)} \end{Bmatrix} \label{eq:flpan}
\end{equation}

The infinite-frequency added mass and linear stiffness coefficients for the FDS are given by Eqs. \eqref{eq:added} and \eqref{eq:stiff}, respectively. The coefficients are computed by PANShip, though they may be computed using any linear, frequency-domain numerical tool.

\footnotesize
\begin{align}
\mathbf{A} &= \begin{bmatrix} 6.9e3 & 0 & 1.2e4 & 0 & 3.3e6 & 0 \\
					0 & 6.0e6 & 0 & -4.7e5 & 0 & 6.5e6 \\
					1.1e4 & 0 & 4.3e6 & 0 & 3.2e7 & 0 \\
					0 & -4.4e5 & 0 & 7.7e6 & 0 & 3.6e7 \\
					3.3e6 & 0 & 3.2e7 & 0 & 2.2e9 & 0 \\
					0 & 6.5e6 & 0 & 3.6e7 & 0 & 4.3e8 \end{bmatrix} \label{eq:added} \\		
\mathbf{C} &= \begin{bmatrix} 0 & 0 & 0 & 0 & 0 & 0 \\
					0 & 0 & 0 & 0 & 0 & 0 \\
					0 & 0 & 1.01e7 & 0 & 3.65e7 & 0 \\
					0 & 0 & 0 & 4.49e7 & 0 & 0 \\
					0 & 0 & 3.65e7 & 0 & 6.25e9 & 0 \\
					0 & 0 & 0 & 0 & 0 & 0 \end{bmatrix} \label{eq:stiff}
\end{align}
\normalsize

Linearized damping coefficients are not produced by PANShip, so a fraction of the critical damping is used where restoring forces are present, i.e. heave, roll, and pitch modes, as given by Eq. \eqref{eq:critdamp}, where $\alpha$ is taken as 0.1.

\begin{equation}
B_{ii} = \alpha 2 \sqrt{C_{ii} (M_{ii}+A_{ii})}, i=3,4,5 \label{eq:critdamp}
\end{equation}

Since the wave conditions under consideration are head-seas ($\mu$~=~180$^{\circ}$), Eqs. \eqref{eq:transpan} and \eqref{eq:rotpan} and Eqs. \eqref{eq:traneomcorr} - \eqref{eq:flpan} will effectively reduce to 3-DOF systems, where only heave, pitch, and surge are excited. In the simulations, the surge velocity $\dot{x}_{s}$ is also held to a fixed value at 35.4~knots (18.2~m/s).

\subsection{Characterization of a Single Seaway}\label{sect:singlesea}

First, the performance of the method is evaluated in a single seaway, especially to understand the optimal stencil length $k$ for the data set. In contrast to Section \ref{sect:duff}, the PANShip data may include hydrodynamic memory effects, which means that the forces and moments at time $t$~=~$t_{n}$, for example, do not depend only on the state at $t_{n}$, but rather on prior states as well. Furthermore, the size of the training data set is investigated.

\begin{figure*}[t]
	\centering
		\includegraphics{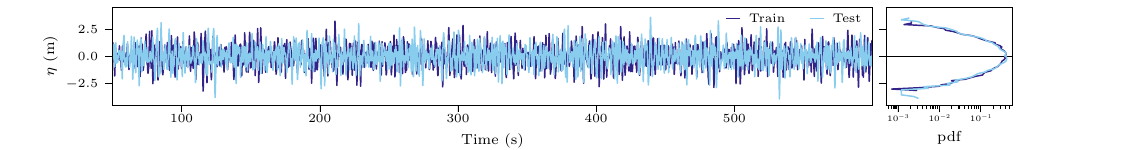}
	  \caption{Training and testing waves for a single seaway: $H_{s}$~=~4~m, $T_{p}$~=~7.5~s, $N_{ZUC}$~=~293. The wave elevation is one of the input features of the ML model.}\label{fig:trainpanwaves}
\end{figure*}

\begin{figure*}[H]
	\centering
		\includegraphics{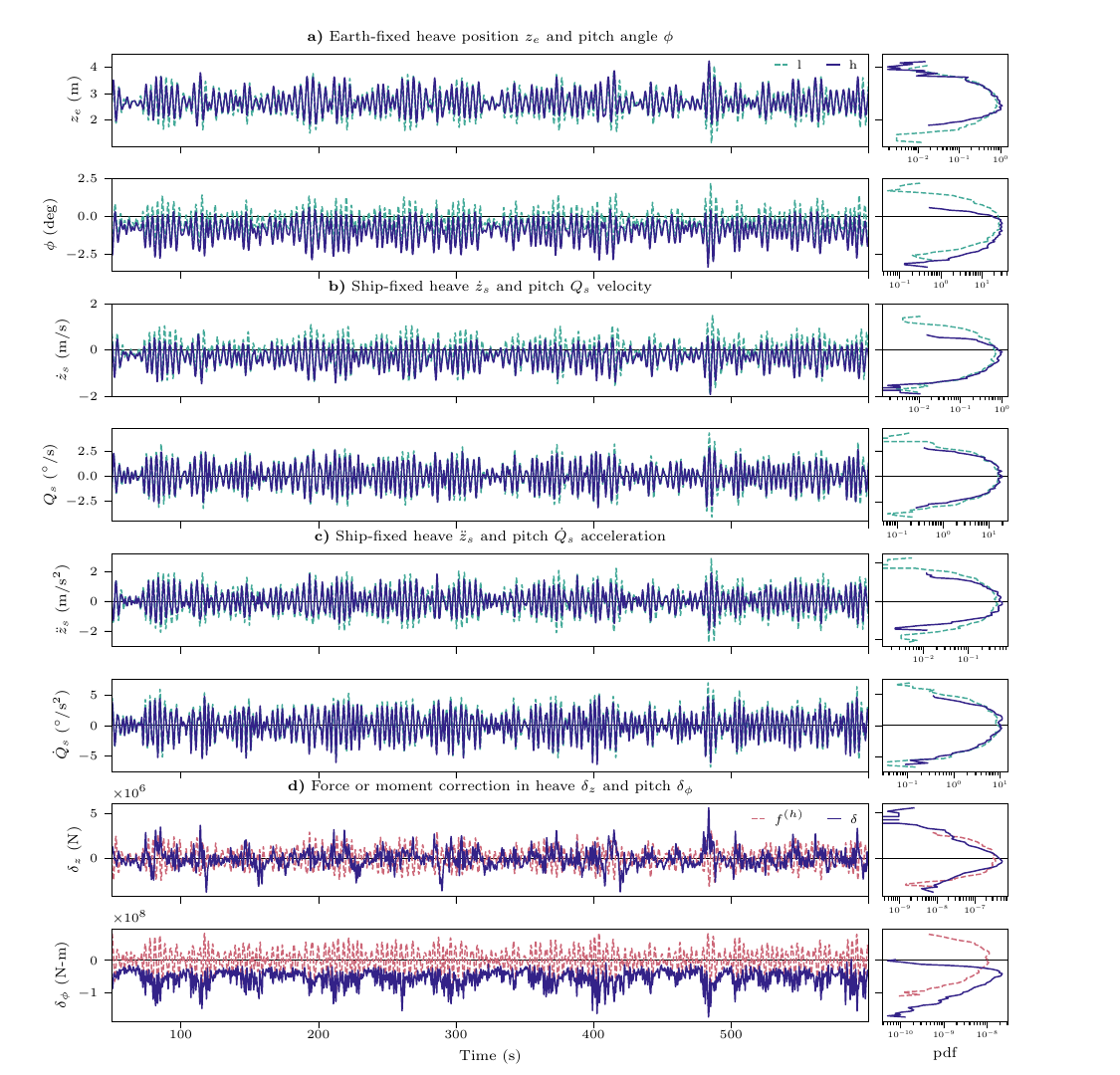}
	  \caption{Training data for the FDS in a single seaway: $H_{s}$~=~4~m, $T_{p}$~=~7.5~s. \textbf{(a) - (c)} comprise the responses and show both the high-fidelity $(h)$ and low-fidelity $(l)$ levels from Table \ref{tab:pannl}. These data are input features of the ML model. \textbf{(e)} comprises force and moment correction and the total high-fidelity force or moment $f^{(h)}$ is also included for comparison. These data are the output of the ML model.}\label{fig:trainpanacc}
\end{figure*}

The low- and high-fidelity responses from Table \ref{tab:pannl} are computed in a single seaway of $H_{s}$~=~4~m and $T_{p}$~=~7.5~s, $N_{ZUC}$~=~293. Two different realizations of the seaway are prepared, each with a different set of random component phase angles, as shown by Figure \ref{fig:trainpanwaves}. The first wave realization is used for training and the second used for testing. All data is sampled to a timestep of $\Delta t$~=~0.1~s. The time series and pdfs of the earth-fixed heave $z_{e}$, Euler angle for pitch $\theta$, ship-fixed heave and pitch velocities $\dot{z}_{s}$, $Q_{s}$, and ship-fixed heave and pitch accelerations $\ddot{z}_{s}$, $\dot{Q}_{s}$ are shown in panels a) through c), respectively, in Figure \ref{fig:trainpanacc}. Both the low- and high-fidelity data are included, in part to show visually the inaccuracies associated with the low-fidelity responses, which are most evident in the tails of the distributions.

$\boldsymbol{\delta}$ is also computed for the training waves and is shown in panel d) of Figure \ref{fig:trainpanacc} for heave and pitch. Also included in the figure are the low-fidelity forces $\mathbf{f}^{(l)}$ and high-fidelity forces $\mathbf{f}^{(h)}$ for comparison. 

The training wave elevation from Figure \ref{fig:trainpanwaves} and the responses and force correction in Figure \ref{fig:trainpanacc} comprise the entire training data set which is used to train the ML model.

\subsubsection{Stencil Size $k$}\label{sect:stencil}

Prior work in \cite{marlantes2023} shows that a small $k$ is effective for nonlinear forces which depend primarily on the instantaneous state, however, larger $k$ is needed for forces which include memory. 

To determine the optimal value for the stencil length $k$ in this work, models are trained for $\boldsymbol{\delta}$ using the full training data set in Figures \ref{fig:trainpanwaves} and \ref{fig:trainpanacc}, for $k$ ranging from 1 to 100. The first 50~s of the response data are omitted from the training data set as this period corresponds to a transient region. The timestep $\Delta t$ is 0.1~s, so the stencil covers a time period of 0.1~s to 10~s, or about 0.013$T_{p}$ to 1.33$T_{p}$. The trained models are used to make predictions of the responses in the full 600~s testing waves, and the $L_{2}$, $L_{\infty}$, and $\mli{JSD}$ metrics are computed over the last 500~s of the predicted time-series.

\begin{figure}
	\centering
		\includegraphics{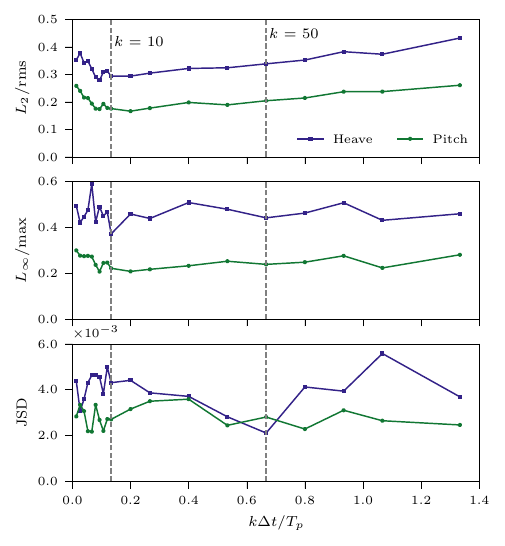}
	  \caption{$L_{2}$, $L_{\infty}$, and $\mli{JSD}$ prediction errors in testing waves $H_{s}$~=~4~m, $T_{p}$~=~7.5~s, for a range of stencil lengths, $k$~=~$[1~:~100]$, $\Delta t$~=~0.1~s.}\label{fig:errk}
\end{figure}

Figure \ref{fig:errk} gives the prediction errors over the range of stencil lengths, given as a ratio relative to the dominant wave period $T_{p}$. Only errors for the ship-fixed heave and pitch acceleration $\ddot{z}_{s}$ and $\dot{Q}_{s}$ are shown, as prediction errors in the velocity and position tend to be similar, if not better, due to the filtering effect of the integration process. 

The $L_{2}$ error and $L_{\infty}$ errors are minimized when $k$ is about 10 ($0.13T_{p}$). However, the $\mli{JSD}$ is minimized, especially for the heave responses, when $k$ is about 50 ($0.67T_{p}$). Because the $\mli{JSD}$ gives an indication of the performance of the model when predicting the tails of the response distribution, a $k$ of 50 is expected to minimize errors in the extreme values. However, a larger $k$ will also lead to higher computational expense.

\subsubsection{Training Data Size}\label{sect:trainsize}

\begin{figure}
	\centering
		\includegraphics{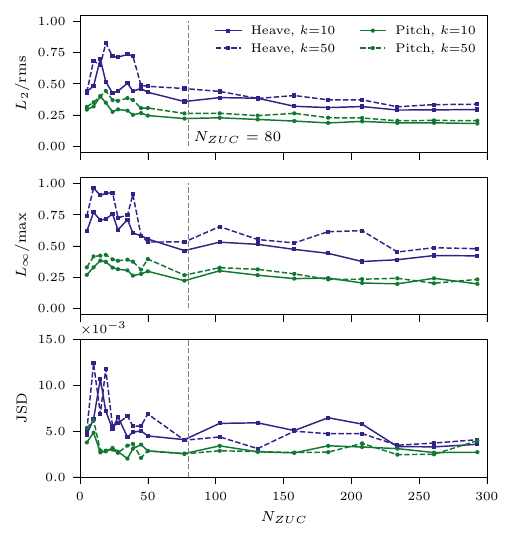}
	  \caption{$L_{2}$, $L_{\infty}$, and $\mli{JSD}$ prediction errors vs training data set size measured in the number of wave Zero-Up-Crossings $N_{ZUC}$. The testing waves are $H_{s}$~=~4~m, $T_{p}$~=~7.5~s, sampled at $\Delta t$~=~0.1~s. Two stencil lengths $k$~=~10 and 50 are used to configure the ML model for comparison.}\label{fig:errtsize}
\end{figure}

A ML model for $\boldsymbol{\delta}$ is trained for increasing portions of the training data shown in Section \ref{sect:stencil} and Figures \ref{fig:trainpanwaves} and \ref{fig:trainpanacc}, starting at $N_{ZUC}$~=~5 (10~s total time) up to the full realization including all $N_{ZUC}$~=~293 (550~s total time). Values of $k$~=~10 and 50, selected from Figure \ref{fig:errk}, are used to configure the models. The first 50~s of the response data are omitted from the training data set as this period corresponds to a transient region. Each trained model is used to predict the responses in the testing waves, and the $L_{2}$, $L_{\infty}$ and $\mli{JSD}$ metrics are computed for each model. Figure \ref{fig:errtsize} shows the prediction errors over increasing training dataset size.

Figure \ref{fig:errtsize} shows that the performance of the model begins to converge at or above approximately $N_{ZUC}$~=~80. This convergence is similar to what is observed in Section \ref{sect:duffsize} for the Duffing equation, as well as the findings in \cite{marlantes2022}. A training dataset encompassing $N_{ZUC}$~=~107 will be used for the remaining results in this section.

\subsection{Generalizability in Different Seaways}\label{sect:transfds}

Both low- and high-fidelity responses from Table \ref{tab:pannl} are computed in each of the nine wave conditions shown in Figure \ref{fig:specs} with total time series lengths of 600~s ($N_{ZUC}$~$\approx$~250). However, only a single moderate wave condition, corresponding to $H_{s}$~=~4.0~m and $T_{p}$~=~8.5~s, is used for training data. Following the results in Figure \ref{fig:errtsize}, the training data is further restricted to only the first 150~s ($N_{ZUC}$~=~107) of the time series. Feed-forward, densely-connected neural networks are configured with three hidden layers, 30 cells per layer, ReLU activation functions (ref. Table~\ref{tab:ncml}), and $k$~=~10 are trained for the heave force correction $\delta_{z}$ and pitch moment correction $\delta_{\theta}$. In addition, an LSTM model configured with 2 hidden layers, and 50 cells per layer (ref. Table~\ref{tab:lstmml}) is also trained on the same training dataset. The trained models are used to make predictions of the earth- and body-fixed heave and pitch responses in the entire set of nine wave conditions. 

\begin{figure*}[t]
     \centering
     \begin{subfigure}[b]{1.0\textwidth}
         \includegraphics{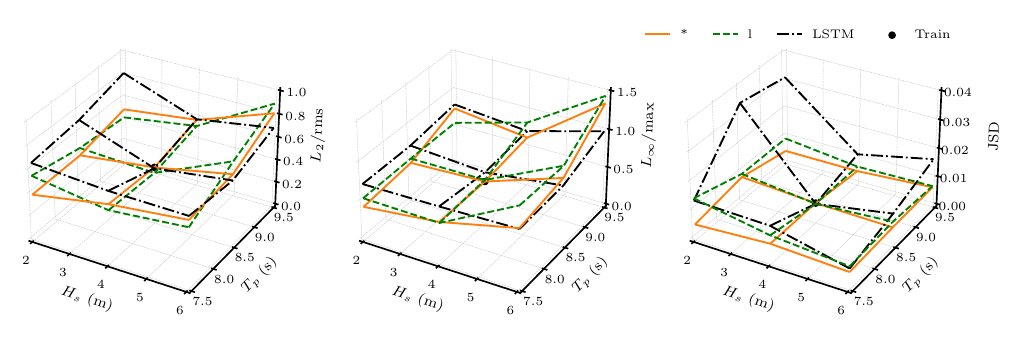}
	  \caption{Heave acceleration $\ddot{z}_{s}$.}\label{fig:heaveaccerr}
     \end{subfigure}
     \vfill
     \begin{subfigure}[b]{1.0\textwidth}
         \includegraphics{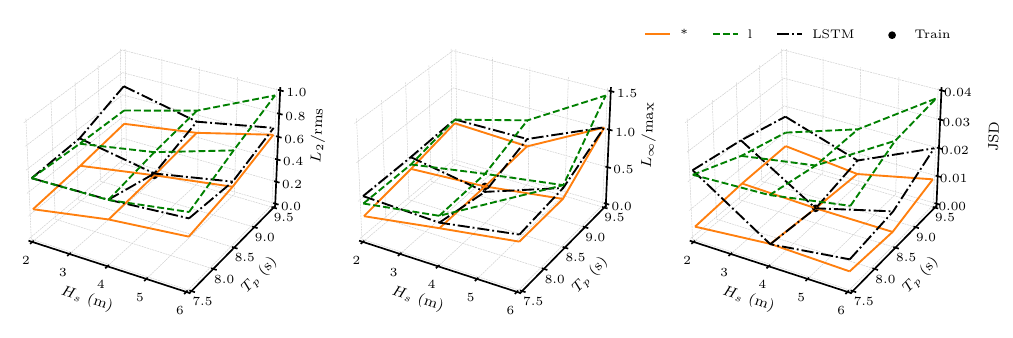}
	  \caption{Pitch acceleration $\dot{Q}_{s}$.}\label{fig:pitchaccerr}
     \end{subfigure}
        \caption{Ship-fixed acceleration prediction $L_{2}$, $L_{\infty}$, and $\mli{JSD}$ errors in different seaways.}
        \label{fig:accerr}
\end{figure*}

The $L_{2}$, $L_{\infty}$ and $\mli{JSD}$ error metrics for the predictions are computed relative to the known high-fidelity responses $(h)$, and the results for the ship-fixed heave $\ddot{z}_{s}$ and pitch $\dot{Q}_{s}$ acceleration are shown in Figure \ref{fig:heaveaccerr} and \ref{fig:pitchaccerr}, respectively. Performance metrics for the low-fidelity model $(l)$ and LSTM predictions are included for comparison. The hybrid method performs similarly to the LSTM in terms of $L_{2}$ and $L_{\infty}$ norms, though improved performance is found in sea-states with smaller $H_{s}$ where the $L_{2}$ error for the LSTM prediction grows. The hybrid method yields improved $\mli{JSD}$ in all wave conditions, especially in the pitch responses, where a significant improvement beyond the low-fidelity benchmark is attained. The $\mli{JSD}$ for the hybrid method predictions is also more consistent in magnitude across the set of wave conditions, where the LSTM predictions tend to vary considerably. 

\begin{figure*}[t]
     \centering
     \begin{subfigure}[b]{1.0\textwidth}
         \includegraphics{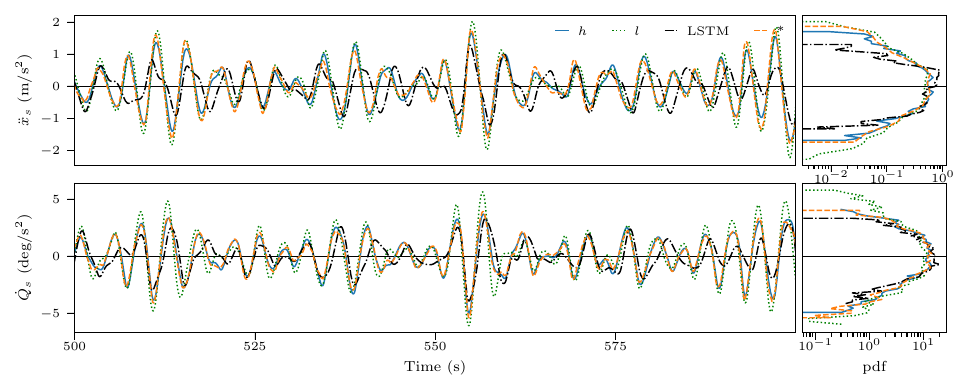}
	  \caption{$H_{s}$~=~2~m, $T_{p}$~=~9.5~s.}\label{fig:accts1}
     \end{subfigure}
     \vfill
     \begin{subfigure}[b]{1.0\textwidth}
         \includegraphics{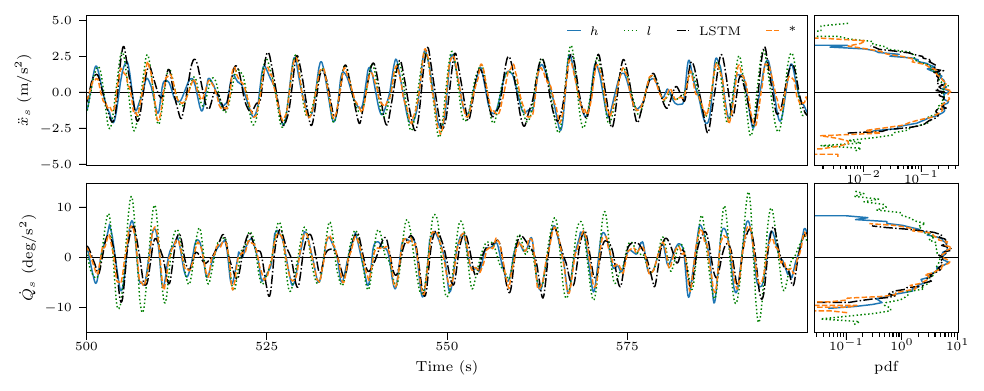}
	  \caption{$H_{s}$~=~6~m, $T_{p}$~=~7.5~s.}\label{fig:accts2}
     \end{subfigure}
        \caption{Ship-fixed heave $\ddot{z}_{s}$ and pitch $\dot{Q}_{s}$ acceleration time-series predictions in irregular waves.}
        \label{fig:accts}
\end{figure*}

Figures \ref{fig:accts1} and \ref{fig:accts2} show the corresponding ship-fixed $\ddot{z}_{s}$ and pitch $\dot{Q}_{s}$ acceleration time series predictions for the least steep $H_{s}$~=~2.0~m, $T_{p}$~=~9.5~s, and most steep $H_{s}$~=~6.0~m, $T_{p}$~=~7.5~s, seaways, respectively. Also shown are the pdfs of the response, which encompass the last 500~s of time series. The two examples are selected becuase they are seaways in the test dataset which are furthest from the training waves. It is also expected that the low-fidelity model $(l)$, which is linear per Table \ref{tab:pannl}, will perform best in the longest period, smallest-$H_{s}$ wave conditions where the data-driven models may have little to gain, but the short, highest-$H_{s}$ waves will lead to the greatest improvement over the low-fidelity model. However, the complete set of predictions for all nine wave conditions can be found in Appendix \ref{app:addfds}.

Figure \ref{fig:accts1} shows that the method offers improved performance over the LSTM model in the lower sea-state, especially in the heave response predictions, where the LSTM model introduces a large phase error. The hybrid method also predicts the tails of the response distribution more accurately than the LSTM model, which underpredicts. The LSTM model demonstrates better performance for the pitch responses, though the predicted distribution is truncated. The hybrid method predictions for pitch acceleration are in reasonable agreement with the high-fidelity reference data. 

In the larger-$H_{s}$ seaway shown in Figure \ref{fig:accts2}, the performance of the hybrid method and LSTM model are more similar. The hybrid method yields slightly better predictions in heave, yet both models perform similarly in pitch. This suggests that the LSTM model can transfer to larger seaways better than smaller seaways.

In terms of computational cost during inference, the LSTM model is similar to the hybrid method, unless small timesteps are used. The hybrid method must make a model inference on every timestep, so timestep size has a greater influence on computational cost than for the LSTM method. During training, the LSTM model takes signficantly more time to train, on the order of hours, where the simple neural networks in the hybrid method take only minutes to train.

In general, the features of the predictions in the PANShip data are consistent with the Duffing equation example in Section \ref{sect:transduff}. The added physics in the hybrid method help to enforce linearity in low-$H_{s}$ seaways, yet the data-driven force correction allows for improved accuracy in larger, more nonlinear, seaways. The LSTM model performs quite well, but struggles to exhibit the same level of generalizability in different wave conditions, especially when evaluated based on the quality of the predicted response distributions. The hybrid method results also appear to be bounded by the linear physics. This is a qualitative statement, but can be observed in the complete time-series results for the FDS in Appendix~\ref{app:addfds}. 

However, if the training dataset size is increased beyond the minimal $N_{ZUC}$~=~107 size, to include the entire available time-series for example, the performance of the LSTM and hybrid methods become more similar. This is expected, as with adequate training data the LSTM model can be quite effective. However, the virtues of the hybrid method are a reduced training dataset while maintaining generalizability across wave conditions. 

\section{Conclusions}\label{sect:concl}

This work applies a hybrid force-correcting machine learning method to two case studies: predicting the nonlinear responses of a Duffing equation subject to irregular excitation, and predicting the high-fidelity heave and pitch responses of a Fast Displacement Ship (FDS) in irregular head seas. In both cases, the domain generalizability of the method is considered by training the model in one seaway, and using the trained model to make predictions in other seaways, and computing performance metrics relative to known reference data. The results are also benchmarked against predictions from a purely data-driven LSTM model.

In the Duffing study, irregular waves are modeled using a generic Bretschnieder-form energy spectrum, with wave heights ranging from 0.01 up to 1.5~m, with training waves of $H_{s}$~=~1.0~m, all with a peak frequency $\omega_{p}$ of 1.0~rad/s. Furthermore, multiple low-fidelity forcing models are considered, including different linear terms, such as linear wave excitation, linear restoring forces, and linear damping forces, and the influence of the choice of low-fidelity forcing model is demonstrated. The minimum size of the training dataset, in terms of the number of wave Zero-Up-Crossings $N_{ZUC}$ is investigated.

In the FDS case, nine irregular wave conditions are considerd, with $H_{s}$~=~2, 4, 6~m, and $T_{p}$~=~7.5, 8.5, 9.5~s, all modeled using a JONSWAP energy spectrum, with $H_{s}$~=~4~m, $T_{p}$~=~8.5~s used for training. The response data are computed at two levels of fidelity, a low-fidelity $(l)$ model which is based on linear theory, and a high-fidelity $(h)$ model which uses second-order direct pressure integration, both afforded by the boundary element code, PANShip. The stencil length $k$ and the training dataset size are characterized for a single seaway.

The following conclusions may be drawn:

\begin{itemize}
\item Including additional linear physics in the low-fidelity forcing model $f^{(l)}$ improves generalizability.
\item The method requires a training dataset comprising between 50 and 100 Zero-Up-Crossings $N_{ZUC}$ in the irregular wave record. 
\item A stencil length $k\Delta t$ of 10-20\% of the dominant wave period is sufficient.
\item Predictions of response distributions, especially in the tails, are more consistent when linear physics are included, vs. a purely data-driven approach.
\end{itemize}  

Perhaps the most significant conclusion from this work is that linear physics, combined with small and simple neural networks, can outperform much more sophisticated machine learning models such as LSTM neural networks when only a small training dataset is available. For marine dynamics applications, where training data may be severely limited due to cost, the proposed method may be advantageous. 

\section{Acknowledgements}
The authors would like to gratefully acknowledge the support from the Office of Naval Research (ONR) Grant Number N00014-22-1-2509 and the University of Michigan Rackham Merit Fellowship (RMF).

 \appendix
 
\section{Eigenvalue Analysis}\label{app:eig}
The low-fidelity model is considered analytically to gain a better understanding of how the system will behave in the limit when $\delta \to 0$. This is important for numerical reasons, but also provides some insight into the system dynamics.

Consider the single degree-of-freedom system given by Eq. \eqref{eq:sdofeom}, which includes the physical mass $m$, linear added mass $a_{1}$, hydrodynamic damping $b_{1}$, and hydrostatic restoring $c_{1}$ terms. The right-hand-side includes the low-fidelity wave excitation force $f_{w}$ and the ML-driven force correction $\delta$. This form of the equation of motion corresponds closely to the low-fidelity forcing model A from Table \ref{tab:dufffl} in Section \ref{sect:duff}, where the most linear physics are retained.

\begin{equation}
(m+a_{1})\ddot{\xi} + b_{1}\dot{\xi} + c_{1}\xi = f_{w} + \delta \label{eq:sdofeom}
\end{equation} 

Substituting $v_{1}$~=~$\xi$ and $v_{2}$~=~$\dot{\xi}$, the system is reduced to standard first-order form in Eq. \eqref{eq:sdofsf}.

\begin{equation}
\underbrace{\begin{Bmatrix} \dot{v}_{1} \\ \dot{v}_{2} \end{Bmatrix}}_{\mathbf{\dot{v}}} = \underbrace{\begin{bmatrix} 0 & 1 \\ \frac{-c_{1}}{m+a_{1}} & \frac{-b_{1}}{m+a_{1}} \end{bmatrix}}_{\mathbf{Q}} \underbrace{\begin{Bmatrix} v_{1} \\ v_{2} \end{Bmatrix}}_{\mathbf{v}} + \underbrace{\begin{Bmatrix} 0 \\ \frac{f_{w}+\delta}{m+a_{1}} \end{Bmatrix}}_{\mathbf{q}} \label{eq:sdofsf}
\end{equation}

The eigenvalues $\lambda_{1,2}$ of matrix $\mathbf{Q}$ determine the linear stability and fixed point behavior of the system. Refering back to Table \ref{tab:dufffl} in Section \ref{sect:duff}, different choices of low-fidelity forcing models may include some, or all, of the linear terms shown in Eq. {eq:sdofeom}. The eigenvalues are computed for each model in Table \ref{tab:dufffl} and are shown in Table \ref{tab:eigs}. Note that $\omega_{n}$ is the natural frequency of the system.

\begin{table}
\caption{Eigenvalues for different low-fidelity forcing models.}\label{tab:eigs}
\begin{tabular*}{\tblwidth}{ R R } 
\toprule
Models A and B & $\lambda_{1,2} = \frac{b_{1}}{2(m+a_{1})} \pm \frac{\sqrt{b_{1}^{2} - 4c_{1}(m+a_{1})}}{2(m+a_{1})}$ \\
\midrule
Models C and D & $\lambda_{1,2} = \pm \sqrt{-\frac{c_{1}}{m+a_{1}}}=\pm i\omega_{n}$ \\
\midrule
Model E & $\lambda_{1,2} = 0$ \\
\bottomrule
\end{tabular*}
\end{table}

The eigenvalues for models A and B will always have a real part when $b_{1} > b_{c}$, where $b_{c}$ is the critical damping coefficient given by $2\sqrt{c_{1}(m+a_{1})}$. Therefore, with sufficient choice of $b_{1}$, Eq. \eqref{eq:sdofeom} can be solved using an explicit numerical integration scheme provided $\Delta t$ is small, though this may not always be preferable. Furthermore, including added mass in the low-fidelity forcing model actually increases $b_{c}$ thereby requiring larger $b_{1}$ to maintain  numerical stability. 

Models C, D, and E will be unconditionally unstable for explicit numerical schemes for all $\Delta t > 0$ because $\lambda_{1,2}$ are purely imaginary or, in the case of model E, zero. Therefore, implicit time integration is preferred for solving hybrid ODEs such as Eq. \eqref{eq:sdofeom}, and all results given in this paper utilize an implicit second-order backward difference scheme as outlined in Appendix~\ref{app:impl}

\begin{figure}
	\centering
		\includegraphics{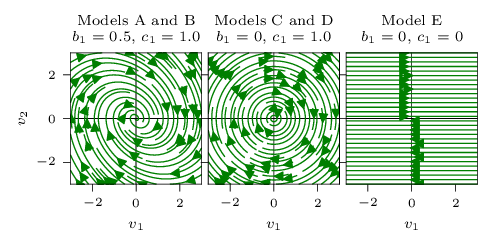}
	  \caption{Phase portraits of Eq. \eqref{eq:sdofeom} subject to different low-fidelity forcing models in Table \ref{tab:dufffl}.}\label{fig:phase}
\end{figure}

The eigenvalues shown in Table \ref{tab:eigs} also shed light on the fixed point behvaior of the low-fidelity system as $\delta_{i} \to 0$, or in other words the linear limit. This is part of why the hybrid model is able to retain accuracy in low-$H_{s}$ seaways. Figure \ref{fig:phase} gives the phase portraits for the low-fidelity models A - E, grouped by their eigenvalues. It is clear that models A and B have a fixed point at $v_{1}$~=~$v_{2}$~=~0, which is a spiral because $\mli{Im}(\lambda_{1,2}) \neq 0$. Models C and D also have a fixed point at $v_{1}$~=~$v_{2}$~=~0, however, it is a center due to the lack of damping. Model E is interesting, because fixed points span the entire plane. This means that the stable solution will depend entirely on $\delta_{i}$, which makes it prone to prediction errors (recall the predicted $\delta^{*}$ is subject to some error $\epsilon$). The result of this sensitivity, and lack of fixed points, is clearly shown in the prediction errors in low-$H_{s}$ in Figure \ref{fig:errhs} in Section \ref{sect:duff}.

\section{Implementation Details}\label{app:impl}

The Eqs. \eqref{eq:traneomcorr} and \eqref{eq:roteomcorr} from Section \ref{sect:fds} are the 6-DOF ship-fixed equations of motion. It is easiest to solve the equations in the ship-fixed system, however, the earth-fixed position and Euler angles must also be updated in time. Furthermore, reducing the ship-fixed equations to a system of first-order differential equations leads to the 12-element state vector $\mathbf{v}$~=~$\{x_{e},y_{e},z_{e},\phi,\theta,\psi,\dot{x}_{s},\dot{y}_{s},\dot{z}_{s},P_{s},Q_{s},R_{s}\}^{T}$ and rate vector $\mathbf{\dot{v}}$~=~$\{\dot{x}_{e},\dot{y}_{e},\dot{z}_{e},P_{e},Q_{e},R_{e},\ddot{x}_{s},\ddot{y}_{s},\ddot{z}_{s},\dot{P}_{s},\dot{Q}_{s},\dot{R}_{s}\}^{T}$ where the subscript $s$ indicates the ship-fixed frame and the subscript $e$ indicates the earth-fixed frame. The first-order system is given by Eq. \eqref{eq:redform}, where $\boldsymbol{\mathbb{I}}_{6x6}$ is the 6x6 identity matrix, $\mathbf{M}$, $\mathbf{A}$, $\mathbf{B}$, and $\mathbf{C}$ are the physical mass, linear hydrodynamic added mass, linear hydrodynamic damping, and linear hydrostatic restoring matrices outlined in Section \ref{sect:fds}. The external forcing vector $\mathbf{F}$ is composed of the low-fidelity wave excitation forcing model as well as the ML-driven force correction terms $\boldsymbol{\delta}$, both 6x1 vectors.

\begin{align}
\mathbf{\dot{v}} &= \underbrace{\boldsymbol{\mathbf{I}}^{-1}\mathbf{G}}_{\mathbf{Q}}\cdot\mathbf{v}+\underbrace{\mathbf{I}^{-1}\mathbf{F}}_{\mathbf{q}} \label{eq:redform} \\
\mathbf{I} &= \begin{bmatrix} \mathbb{I}_{6x6} & \mathbf{0}_{6x6} \\ \mathbf{0}_{6x6} & \mathbf{M}+\mathbf{A} \end{bmatrix} \\
\mathbf{G} &= \begin{bmatrix} \mathbf{0}_{6x6} & \mathbf{T} \\ -\mathbf{C} & -\mathbf{B} \end{bmatrix} \\
\mathbf{F} &= \{ \mathbf{0}_{6x1},\mathbf{f_{w}^{(l)}}+\boldsymbol{\delta^{*}}\}^{T}
\end{align}

The earth-fixed velocities are related to the ship-fixed velocities by a transformation which depends on the Euler angles, where $\mathbf{T}$ is the 6x6 block-diagonal transformation matrix composed of the translational and rotational transformation matrices given by Eqs. \eqref{eq:t1} and \eqref{eq:t2}, respectively. Note that the subscripts $c$, $s$, and $t$ are used to indicate the cosine, sine, and tangent of the respective Euler angle. 

{\small
\begin{align}
\mathbf{T} &= \begin{bmatrix} \mathbf{T_{1}} & \mathbf{0}_{3x3} \\ \mathbf{0}_{3x3} & \mathbf{T_{2}} \end{bmatrix} \\
\mathbf{T_{1}} &= \begin{bmatrix} \psi_{c}\theta_{c} & -\phi_{c}\psi_{s}+\psi_{c}\phi_{s} & \phi_{s}\psi_{s} + \phi_{c}\psi_{c}\theta_{s} \\ \theta_{c}\psi_{s} & \phi_{c}\psi_{c} + \phi_{s}\psi_{s}\theta_{s} & -\psi_{c}\phi_{s}+\phi_{c}\psi_{s}\theta_{s} \\ -\theta_{s} & \theta_{c}\phi_{s} & \phi_{c}\theta_{c} \end{bmatrix} \label{eq:t1}\\
\mathbf{T_{2}} &= \begin{bmatrix} 1 & \phi_{s}\theta_{t} & \phi_{c}\theta_{t} \\ 0 & \phi_{c} & -\phi_{s} \\ 0 & \frac{\phi_{s}}{\theta_{c}} & \frac{\phi_{c}}{\theta_{c}} \end{bmatrix} \label{eq:t2}
\end{align}
}

In this work, the derivative is discretized using the second-order backward difference scheme (BDF2). The system from Eq. \eqref{eq:redform} discretized using BDF2 is given by Eq. \eqref{eq:bdf2}, where the governing equation is written in terms of the coefficient matrix $\mathbf{Q}$ and the current timestep is $t_{n}$.

\begin{gather}
\frac{3\mathbf{v}_{n+1} - 4\mathbf{v}_{n} + \mathbf{v}_{n-1}}{2\Delta t} = \mathbf{Q}\cdot\mathbf{v}_{n+1} + \mathbf{q}_{n+1} \label{eq:bdf2}
\end{gather}

Because $\boldsymbol{\delta}$ in $\mathbf{q}$ and the embedded Euler angles in $\mathbf{Q}$ make the equations nonlinear, Eq. \eqref{eq:bdf2} cannot simply be reduced to a linear system and solved at each time step. Instead, the system of nonlinear equations $H(\mathbf{v}_{n+1})$ in Eq. \eqref{eq:root2} are solved iteratively using a Newton root-finding technique. This process is outlined by Eqs. \eqref{eq:Nup} and \eqref{eq:jsys}, where the Jacobian $\mathbf{J}$ for the $k$th iterate is approximated numerically using finite-differencing over perturbations to the state variables. The linearized system in Eq. \eqref{eq:jsys} is solved to yield an update vector $\boldsymbol{\alpha}$, which is used to update the state $\mathbf{v}_{n+1}$ to the $k+1$ iterate in Eq. \eqref{eq:Nup}. The root-finding process iterates in $k$ until a vector norm of $\boldsymbol{\alpha}^{k}$ is small.

{\small
\begin{align}
\left[\frac{3}{2\Delta t} \boldsymbol{\mathbb{I}} - \mathbf{Q}\right]\cdot\mathbf{v}_{n+1} - \frac{1}{2\Delta t}\left(4\mathbf{v}_{n} + \mathbf{v}_{n-1}\right) - \mathbf{q}_{n+1} &= \mathbf{0} \label{eq:root} \\
H(\mathbf{v}_{n+1}) &= \mathbf{0} \label{eq:root2}
\end{align}
}

\begin{align}
\mathbf{v}_{n+1}^{k+1} &= \mathbf{v}_{n+1}^{k} + \boldsymbol{\alpha}^{k} \label{eq:Nup}\\
\mathbf{J}^{k}\boldsymbol{\alpha}^{k} &= -H(\mathbf{v}_{n+1}^{k}) \label{eq:jsys}
\end{align}

For the results in this paper, the surge velocity $\dot{x}_{s}$ is held to a fixed value by first specifying the initial condition $\dot{x}_{s}(t=0)$~=~$\dot{x}_{s,0}$ to the desired value and forcing the surge acceleration $\ddot{x}_{s}$ equal to zero before each Newton iteration. A similar approach can also be used to restrain other degrees of freedom.

Here we also have a choice on whether to make the ML-driven update for $\boldsymbol{\delta}$ implicit or explicit. If the update were implicit, the input features to the model, shown in Table \ref{tab:ncml} in Section \ref{sect:ncmethod}, would need to be modified to include the state and wave elevation at $t_{n+1}$, i.e. $\{\xi\}_{n-k}^{n+1}$, $\{\dot{\xi}\}_{n-k}^{n+1}$, $\{\ddot{\xi}\}_{n-k}^{n+1}$ , and $\{\eta\}_{n-k}^{n+1}$. However, testing showed that this resulted in little, if any, improved stability or accuracy and greatly increased computational cost as the ML model had to be evaluated, on average, 2-3x more often during the solution. Therefore, the implementation developed for this paper used an explicit update for $\boldsymbol{\delta}$ which occurs before, but most importantly outside, of the Newton root-finding iteration to avoid redundant model evaluations.

The details in this section also apply to the 1-DOF Duffing equation from Section \ref{sect:duff}. The main difference is the reference frame is inertial, so the state vectors $\mathbf{v}$~=~$\{z_{e},\dot{z}_{e}\}^{T}$ and $\mathbf{\dot{v}}$~=~$\{\dot{z}_{e},\ddot{z}_{e}\}^{T}$, and the matrices $\mathbf{M}$, $\mathbf{A}$, $\mathbf{B}$, and $\mathbf{C}$ will reduce to scalars. The transformation matrix $\mathbf{T}$ will also be 1, so the coefficient matrix $\mathbf{Q}$ will be a 2x2 matrix, and the vector $\mathbf{q}$ will be a 2x1 vector. The Duffing implementation is shared as open-source code in a repository \citep{marlantes2024s}.

\section{Additional FDS Results}\label{app:addfds}

Included in this appendix are the complete heave and pitch time-series predictions from Section \ref{sect:transfds} for the Fast Displacement Ship in the two irregular wave conditions presented in Figure~\ref{fig:accts}, including the earth-fixed heave $z_{e}$ position and pitch Euler angle $\theta$, and the ship-fixed heave $\dot{z}_{s}$ and pitch $Q_{s}$ velocity. Any other state variables follow by transformation. The results in all figures come from models that are trained in 150~s ($N_{ZUC}$~=~107) of data from the moderate seaway $H_{s}$~=~4~m, $T_{p}$~=~8.5~s.

\begin{figure}[h]
	\centering
		\includegraphics{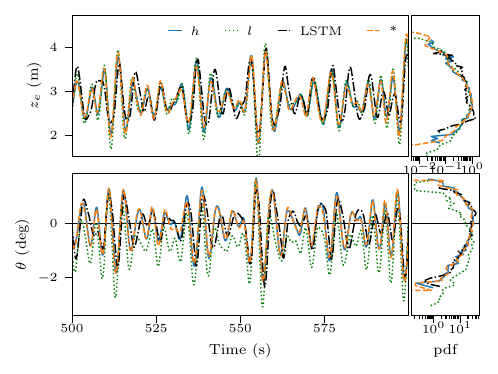}
	  \caption{Heave $z_{e}$ and pitch $\theta$ in $H_{s}$~=~2~m, $T_{p}$~=~9.5~s.}
\end{figure}

\vspace{-12pt}
\begin{figure}[h]
	\centering
		\includegraphics{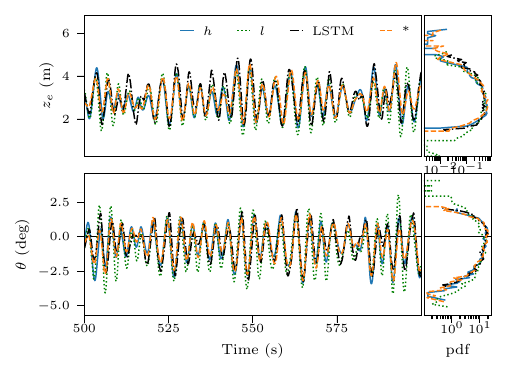}
	  \caption{Heave $z_{e}$ and pitch $\theta$ in $H_{s}$~=~6~m, $T_{p}$~=~7.5~s.}
\end{figure}

\begin{figure}[h]
	\centering
		\includegraphics{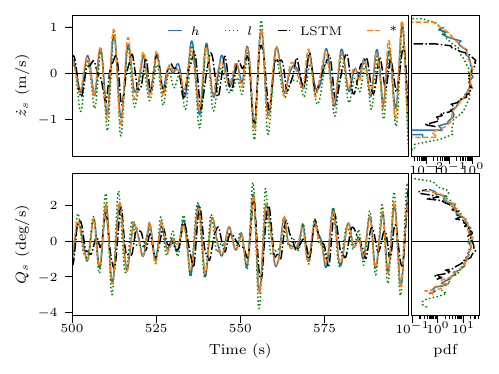}
	  \caption{Heave $\dot{z}_{s}$ and pitch $Q_{s}$ in $H_{s}$~=~2~m, $T_{p}$~=~9.5~s.}
\end{figure}

\begin{figure}[h]
	\centering
		\includegraphics{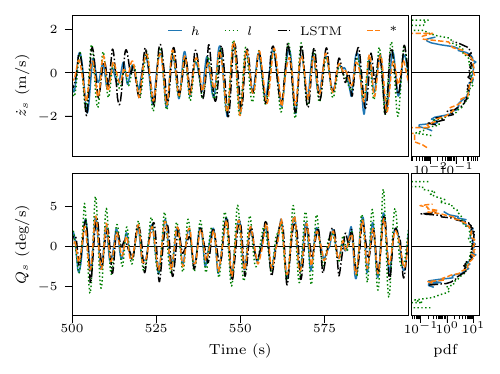}
	  \caption{Heave $\dot{z}_{s}$ and pitch $Q_{s}$ in $H_{s}$~=~6~m, $T_{p}$~=~7.5~s.}
\end{figure}

\printcredits

\bibliographystyle{cas-model2-names}

\bibliography{refs.bib}

\end{document}